\documentclass[runningheads]{llncs}
\usepackage{float}
\usepackage{url}
\usepackage{microtype}
\usepackage[utf8]{inputenc}
\usepackage[T1]{fontenc}
\usepackage{graphicx}
\usepackage{xcolor}
\begin{document}
\title{Defect Segmentation in OCT scans of ceramic parts for non-destructive inspection using deep learning}
\titlerunning{Defect Segmentation in OCT scans of ceramic parts using deep learning}
%
\author{Andrés Laveda-Martínez\inst{1} \and
Natalia P. García-de-la-Puente\inst{1} \and
Fernando García-Torres\inst{1}\and Niels Møller Israelsen\inst{2,3}\and Ole Bang\inst{2,3} \and Dominik Brouczek\inst{4} \and Niels Benson\inst{5}\and Adrián Colomer\inst{1} \and Valery Naranjo\inst{1} }
\authorrunning{A. Laveda-Martínez et al.}
%
\institute{
Instituto Universitario de Investigación en Tecnología Centrada en el Ser Humano, Universitat Politècnica de València, Valencia, Spain
\and Department of Electrical and Photonics Engineering, Technical University of Denmark, Kongens Lyngby, Denmark \and NORBLIS ApS, Virum, Denmark \and Lithoz GmbH, Vienna, Austria \and airCode UG, Duisburg, Germany\\
\email{napegar@upv.es}}

\maketitle              
\begin{abstract}
Non-destructive testing (NDT) is essential in ceramic manufacturing to ensure the quality of components without compromising their integrity. In this context, Optical Coherence Tomography (OCT) enables high-resolution internal imaging, revealing defects such as pores, delaminations, or inclusions. This paper presents an automatic defect detection system based on Deep Learning (DL), trained on OCT images with manually segmented annotations. A neural network based on the U-Net architecture is developed, evaluating multiple experimental configurations to enhance its performance. Post-processing techniques enable both quantitative and qualitative evaluation of the predictions. The system shows an accurate behavior of 0.979 Dice Score, outperforming comparable studies. The inference time of 18.98 seconds per volume supports its viability for detecting inclusions, enabling more efficient, reliable, and automated quality control.

\end{abstract}
\section{Introduction}
Quality control in industrial manufacturing processes is essential to ensure the proper condition of the final product. Currently, most of these processes are carried out manually through visual inspection of the manufactured parts. This approach presents significant limitations related to operator subjectivity, visual fatigue, and the inability to detect certain subsurface defects that are not visible to the naked eye. Therefore, there is a growing need in the industry to develop quality control techniques with real-time monitoring systems that enable early and automatic defect detection during the manufacturing process \cite{YUN2020317}.

To improve this process, Non-Destructive Testing (NDT) methods have gained particular importance in recent years. These techniques are highly valuable for assessing product integrity without causing damage and allow the detection of defects or anomalies in the early stages of production, thus preventing their propagation to later phases. There are numerous NDT methods, among which some stand out as the most commonly used. Firstly, industrial radiography is notable, as it employs short-wave X-rays, gamma rays, or neutrons to penetrate materials \cite{andersson_detection_2016}. Another widely used technique is ultrasonic inspection, which is based on the transmission of mechanical waves to identify internal irregularities \cite{escuderos_ultrasonidos_2010}. The liquid penetrant method is also one of the most common methods for detecting surface defects in dense materials \cite{baldarrago_visual_2015}. Finally, thermography is used to analyse abnormal thermal distributions in materials \cite{maldague2012nondestructive}.

Besides the techniques already described, there are other methods, originally developed for use in fields such as medicine, that are now gaining traction in industrial NDT \cite{fu_progress_2024,su_optical_2012,petersen_non-destructive_2021}. One notable example is Optical Coherence Tomography (OCT), a technique that uses a low-coherence near-infrared light beam ($\sim$800 nm or $\sim$1300 nm) to obtain high-resolution cross-sectional images of both the surface and internal layers of a material \cite{israelsen2019real}. Its ability to acquire three-dimensional information and its high resolution have made OCT widely used in medicine \cite{swanson_oct_2017}. However, due to its potential, it has also begun to be applied in defect detection processes in industrial manufacturing \cite{czajkowski_optical_2010,su_perspectives_2014}. In this work, we focus on its application to ceramic components manufactured using Lithography-based Ceramic Manufacturing (LCM), a process that enables the fabrication of complex, high-resolution ceramic parts. These components are particularly sensitive to internal defects, making them ideal candidates for the application of high-resolution, non-destructive imaging and automated defect detection \cite{lapre_rapid_2024,zorin_mid-infrared_2022}.

Deep Learning (DL) is an advanced technique within the field of Artificial Intelligence (AI), based on deep neural networks, that enables automatic learning of complex patterns from data. Specifically, supervised learning utilises labelled datasets to train models that can predict on new samples. In this context, the combination of DL tools with NDT techniques enables the automatic analysis of images and the detection of defects without damaging the part, aiming to minimise material waste. DL has revolutionised industrial inspection by facilitating automatic detection of defects in images. Specifically, the U-Net architecture has been successfully used for defect detection tasks in industrial settings \cite{wang_segmentation_2023,usamentiaga_automated_2022}. 

The U-Net architecture was initially developed to segment medical images \cite{ronneberger_u-net_2015} and has proven to be very useful in identifying structures or anomalies, such as tumours or cells, in various types of clinical images \cite{three-dimension_epithelial_segmentation}. Its design allows for the combination of fine details with contextual information, making it highly effective for pixel-level segmentation, as demonstrated in the referenced studies. When used in conjunction with imaging techniques such as OCT, this network has successfully segmented internal structures in great detail in medical fields such as ophthalmology \cite{kugelman_comparison_2022,oh_gcn-assisted_2024,morales_retinal_2021,garcia-torres_using_2024} or cancer diagnosis \cite{liu_one-class_2023}. The results obtained in these studies, with Dice Similarity Coefficient values close to 98\% or 99\%, confirm that this combination can automatically detect regions of interest with high precision. 

In the industrial sector, U-net architectures have been applied to detect defects in materials using X-ray or thermographic images \cite{konovalenko_research_2022,jha_deep_2023}. In the context of ceramic parts, U-Net architectures have been used to detect cracks through visual inspection imaging and to segment the 3D microstructure in electron microscopy volumes \cite{hirabayashi_deep_2024,junior_ceramic_2021}. Although the combination of U-Net and OCT has not yet been widely applied in industrial settings, some recent efforts have explored the use of machine learning techniques to analyse OCT data in LCM processes. In particular, Heise et al. \cite{heise_mid-infrared_2024} employed a mid-infrared OCT system alongside a pre-trained ResNet architecture to classify individual B-scans of ceramic components produced by LCM. Their network was able to distinguish between different types of defects, such as voids, inclusions, and contamination. This work highlights the potential of combining OCT imaging and DL for defect detection in additive manufacturing. However, this study was limited to 2D classification and did not address volumetric segmentation.

This study explores the application of a U-Net architecture for defect segmentation in OCT scans of ceramic components manufactured by LCM. Our contribution lies in adapting and validating this approach within an industrial context, demonstrating its potential to improve defect detection accuracy, reduce inspection time, and support the automation of quality control processes in ceramic manufacturing.

\section{Methodology}
The proposed approach for automatic segmentation of subsurface defects in ceramic parts using images obtained by OCT is described in this section. The method is based on the training of convolutional neural networks of type U-Net to identify defective regions in OCT volumes.

\textbf{Problem formulation:}  The problem was formulated as a supervised segmentation task, where each input image was paired with a binary mask representing the locations of internal defects. These image-mask pairs were used to train a neural network. Various model configurations were evaluated to identify the one that most accurately segmented defects in previously unseen images.

\subsection{Generation of binary masks}
Before training the models, it was necessary to generate binary masks through pixel-wise segmentation to train the network. As a first step, the OCT volumes were loaded into MATLAB’s VolumeViewer tool. Manual segmentation was chosen due to the low visibility and lack of homogeneity of the defects present in this type of images, which made their reliable automatic identification difficult.

Several types of defects were considered, but finally, only inclusions were segmented, discarding the rest because they did not present sufficiently clear or consistent visual characteristics to be used for model training. By segmenting a single type of defect, a binary segmentation (defect/no defect) was performed. 

\begin{figure}[ht]
\includegraphics[width=\textwidth]{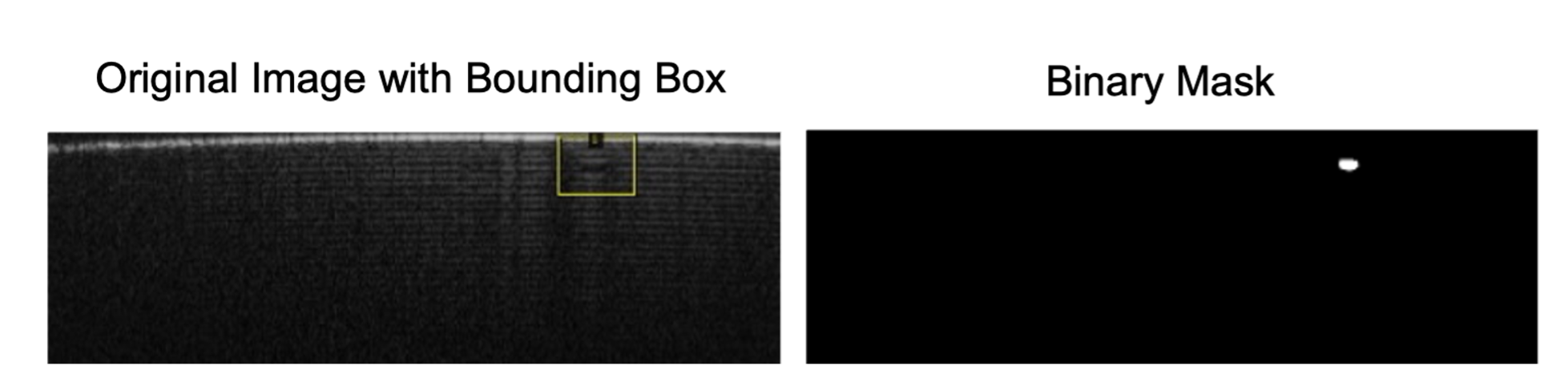}
\caption{Detail of the original OCT image showing the inclusion defect with bounding box and binary mask.}
\label{fig:defects}
\end{figure}

In Figure~\ref{fig:defects}, we can observe the inclusion defect surrounded by a bounding box and its corresponding segmentation. The inclusion typically appears as a small bright region in the OCT image.

\subsection{Preprocessing}
\label{sec:preprocessing}
Once the binary masks were created, and before training the network, it was necessary to preprocess the data by applying transformations to the original images and their masks. Some images initially have a size of 1024 × 700 pixels, others 2048 × 700, but it was observed that the lower part did not contain relevant information for training in any of the images in the entire dataset. 

U-Net networks require input dimensions divisible by 32 to avoid decoding errors. To meet this requirement without losing information, horizontal padding was applied using the Albumentations library, adjusting the width from 700 to 704 pixels, along with two additional transformations. 

Furthermore, a Crop transformation of 352 × 704 was applied to satisfy the divisible-by-32 condition and to crop the image, removing irrelevant parts outside the indicated region. At this stage, the irrelevant lower part was removed. This preprocessing not only allowed for standardising the input sizes, but also prevented common structural errors.

Therefore, once the corresponding masks were created and organised, they were preprocessed along with the images using the indicated transformations and input together with the images into the U-Net neural network for training.

\subsection{U-Net architecture}

For the task of segmenting the defects in the OCT images of ceramic parts, the U-Net architecture was used \cite{garcia2025tfg}. This network is specially designed for segmentation, combining a compression phase that extracts relevant features with an expansion phase that reconstructs the segmentation pixel by pixel. The U-Net architecture is named after the "U" shape of its structure. It is made up of two main paths: a contraction phase (encoder) and an expansion phase (decoder).

\textbf{Contraction phase:}  In this first stage, the network extracts features from the input image by applying consecutive convolutional layers followed by max pooling. This process reduces spatial resolution while capturing the global context of the image.

\textbf{Expansion phase:} The network reconstructs the original resolution of the feature map using upsampling operations and convolutions. At each step, the features of the encoder are combined, allowing the network to recover fine spatial details and improve edge accuracy in the segmentations. Thanks to this combination of encoding and decoding, U-Net can make very accurate predictions even with a small number of training images. 

The implementation was carried out using SMP (Segmentation Models PyTorch). This tool allows loading architectures such as U-Net with different encoders (backbones). In this work, ResNet34 was used as the backbone, loaded with pretrained weights on ImageNet. This means that instead of starting training from scratch with random weights, prior knowledge is leveraged from a network that has already learned to identify basic visual patterns, which is very useful when working with a small dataset.
In this case, a U-Net with an encoder depth parameter set to 5 was used. This indicates that the encoder consists of five main stages or blocks that extract features from the image at different resolutions. Each stage progressively reduces the spatial resolution of the image while increasing the number of channels to capture more complex information. This allows the model to understand patterns that range from very fine details to global structures.

\subsection{Network training and model evaluation}
At this point, we created the binary masks through segmentation, preprocessed the images and masks, and built the neural network. The network was then trained and used to generate predictions.

In training mode, the images used belong to the training and validation sets. First, preprocessing is applied, which in this case consists of a crop. Then, the images and their ground truth (GT) masks are fed into the network. The GT refers to the manually segmented masks that serve as the reference during training. The network generates a prediction that is compared with the GT using a loss function. This loss is used to perform backpropagation, allowing the network’s weights to be updated. Once all images have been processed over multiple epochs, the model with the trained weights is saved.

In evaluation mode, the images used belong to the test set. Just as in training, preprocessing is applied to adjust their size. The images are fed into the network, where the previously trained weights are used. The network outputs an image with per-pixel probability values. A threshold of 0.5 is applied to convert these probabilities into final binary predictions \cite{brownlee_threshold2021}. Once a model has been trained, it is necessary to objectively evaluate its performance. For this purpose, different quantitative metrics were used to evaluate the predictions, which allowed us to measure the performance and compare it with other models.

\section{Experimental Settings}
\subsection{Dataset and evaluation metrics}
The MIR-OCT Scans dataset, available in the Zenodo repository (\url{https://zenodo.org/records/15165755}), contains 30 volumetric OCT scans of ceramic parts in PNG format, including annotations in the form of bounding boxes (BBs), which are coordinates that delimit the defect region and assist in manual segmentation of the volumes.

The volumes consist of between 700 and 1499 slices, with image resolutions of either 1024×700 or 2048×700 pixels, depending on the scan. They include three types of internal defects: pores, delaminations, and inclusions.

To train and evaluate the model, the 30 volumes were divided into three subsets: 21 for training, 6 for validation, and 3 for test. This split allowed the model to learn from the training data and be evaluated on previously unseen images. Each volume corresponds to a single ceramic part. Therefore, the test set includes volumes from parts different from those used in training and validation, ensuring a proper evaluation of the model’s generalization capability. The division is presented in Table~\ref{tab:dataset_split}, which shows the number of volumes used in each subset, as well as the corresponding number of B-scans per set.

\begin{table}[ht]
\caption{Dataset split statistics: number of volumes and B-Scans in each set.}
\label{tab:dataset_split}
\centering
\begin{tabular}{|c|c|c|}
\hline
\textbf{Set} & \textbf{No. of Volumes} & \textbf{No. of B-Scans} \\
\hline
Train      & 21 & 16033 \\
Validation & 6  & 4449 \\
Test       & 3  & 2069 \\
\hline
\end{tabular}
\end{table}

Metrics such as Precision, Recall, and Dice Similarity Coefficient (DSC) were used \cite{mahmoudi2022deep}. The combined use of Precision, Recall, and DSC made it possible to evaluate not only the number of defects detected, but also the accuracy of the segmentations and their overlap with the ground truth.

\subsection{Implementation details}
Fixed hyperparameters were defined for training, such as a batch size of 16, a learning rate of 0.001, and the Adam optimizer. A model checkpointing strategy was implemented based on improvements in the validation loss, combined with early stopping with a patience of 15 epochs. This approach stops training when no improvement is observed, helping to prevent overfitting and reduce computational cost. After training, the final model and its weights were saved. 

To improve performance, five loss configurations were evaluated: BCE, DLS, their equal combination, and two weighted variants (BCE 0.7 + DLS 0.3 and BCE 0.3 + DLS 0.7), as summarized in Table~\ref{tab:loss_configs}.

\begin{table}[ht]
\caption{Summary of training configurations and loss functions used.}
\label{tab:loss_configs}
\centering
\begin{tabular}{|c|l|}
\hline
\textbf{N.º Train} & \textbf{Loss Function} \\
\hline
1 & BCE \\
2 & Dice Loss (DLS) \\
3 & BCE + DLS \\
4 & BCE (0.7) + DLS (0.3) \\
5 & BCE (0.3) + DLS (0.7) \\
\hline
\end{tabular}
\end{table}

Regarding the hardware, we used an NVIDIA DGX system, equipped with 8× NVIDIA A100 GPUs (40 GB each), an AMD EPYC 7742 processor (64 cores, 2.25 GHz, 256 threads), and 1 TB of DDR4 RAM (16×64 GB, 3200 MHz). 
The software used included PyTorch for model training, Albumentations for data augmentation, OpenCV and NumPy for image preprocessing, and Matplotlib for results visualization.

\section{Results}

The validation results for each training configuration are presented in Table~\ref{tab:segmentation_val_results}, showing the performance on the validation set corresponding to each trained model. It can be observed that the first model, trained using the BCE loss function, achieves the best performance metrics during training. To verify the generalization capability of the trained models, they were evaluated on the test set, and the corresponding performance metrics are reported in Table~\ref{tab:segmentation_results}. The reported test metrics correspond to the average values across these volumes.

\begin{table}[ht]
\caption{Segmentation performance metrics for each training configuration in the validation set.}
\label{tab:segmentation_val_results}
\centering
\begin{tabular}{|c|c|c|c|}
\hline
\textbf{N.º Train} & \textbf{DSC} & \textbf{Precision} & \textbf{Recall} \\
\hline
\textbf{1}   & \textbf{0.944} & \textbf{0.967} & 0.923 \\
2   & 0.938 & 0.938 & \textbf{0.937} \\
3   & 0.926 & 0.920 & 0.932 \\
4   & 0.938 & 0.940 & 0.936 \\
5   & 0.914 & 0.898 & 0.930 \\
\hline
\end{tabular}
\end{table}

\begin{table}[ht]
\caption{Segmentation performance metrics for each training configuration in the test set and inference time.}
\label{tab:segmentation_results}
\centering
\begin{tabular}{|c|c|c|c|c|}
\hline
\textbf{N.º Train} & \textbf{DSC} & \textbf{Precision} & \textbf{Recall} & \textbf{Inference (s)} \\
\hline
1   & 0.974 & 0.978 & 0.973 & \textbf{18.526} \\
2   & 0.959 & 0.963 & 0.958 & 18.543 \\
3   & 0.978 & 0.982 & 0.977 & 18.546 \\
\textbf{4} & \textbf{0.979} & \textbf{0.983} & \textbf{0.978} & 18.983 \\
5   & 0.974 & 0.979 & 0.973 & 18.593 \\
\hline
\end{tabular}
\end{table}

Training configuration 4 was selected as the final model due to its superior performance in the test set, achieving the highest values of DSC (0.979), Precision (0.983), and Recall (0.978). 

To demonstrate the effectiveness of our approach, we compared it with a similar study on microvessel segmentation in IVOCT images \cite{Lee2022}, which reported a DSC of 0.73±0.10. In contrast, our model achieved a DSC of 0.979, demonstrating higher segmentation accuracy and robustness in defect detection.

Although the other study primarily focuses on defect detection rather than segmentation, we also mention it because it uses OCT imaging to detect inclusions in ceramic components \cite{Heise2024}, reporting an F1 Score of 0.76. In contrast, our work addresses segmentation-level analysis based on OCT, which is crucial for more detailed industrial inspection. This further highlights the importance of OCT-based segmentation analysis for industrial applications, especially given the superior results achieved by our method.

Our model processes volumes of 1024 × 700 × 700 in 18.98 seconds, with an average time of 0.027 seconds per slice. In comparison, a similar study on OCT-based cancer organ segmentation \cite{diagnostics14121217} uses volumes of 1200 x 800 x 698, reporting an inference time of 30 seconds per volume and 0.043 seconds per slice. This indicates that our method is more efficient both in processing the entire volume and on a per-image basis.

\begin{figure} [ht]
\includegraphics[width=\textwidth]{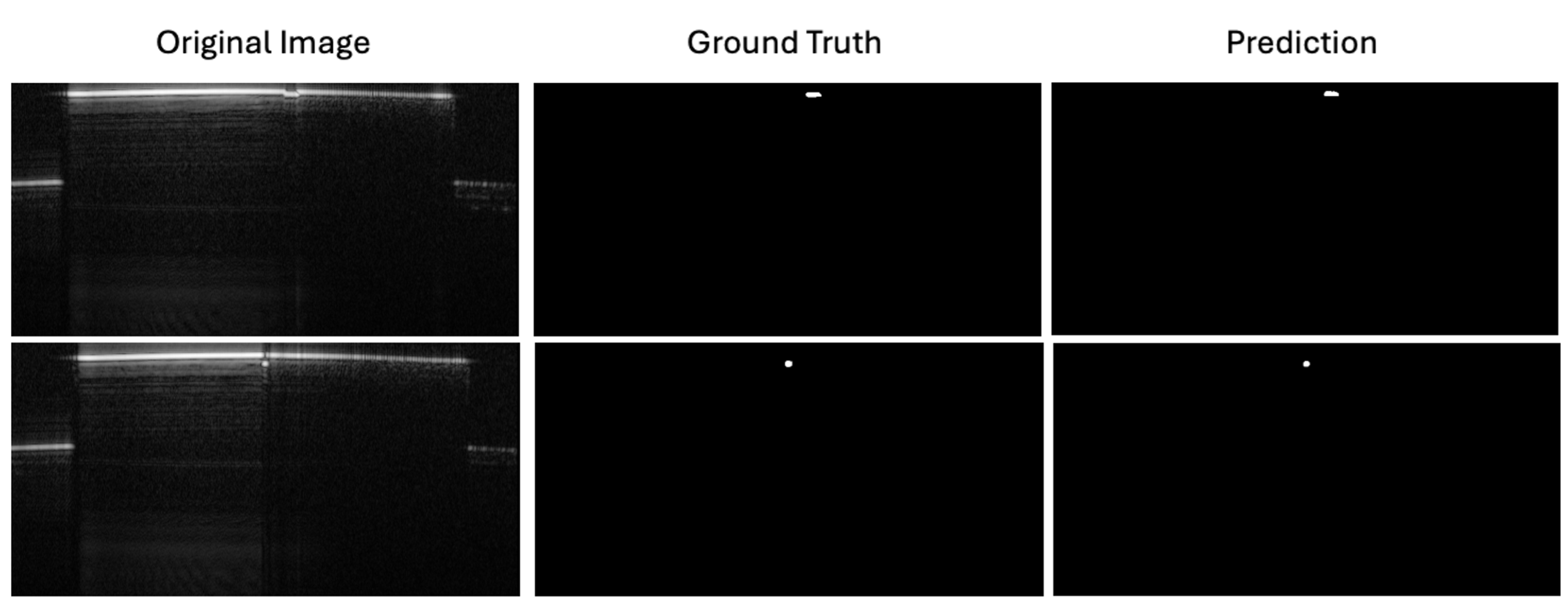}
\caption{Representative segmentation results (Prediction) compared with manual annotations (Ground Truth). The model consistently achieves accurate defect localization, showing strong concordance with expert labels.}
\label{fig:cuali}
\end{figure}

\begin{figure} [ht]
\includegraphics[width=\textwidth]{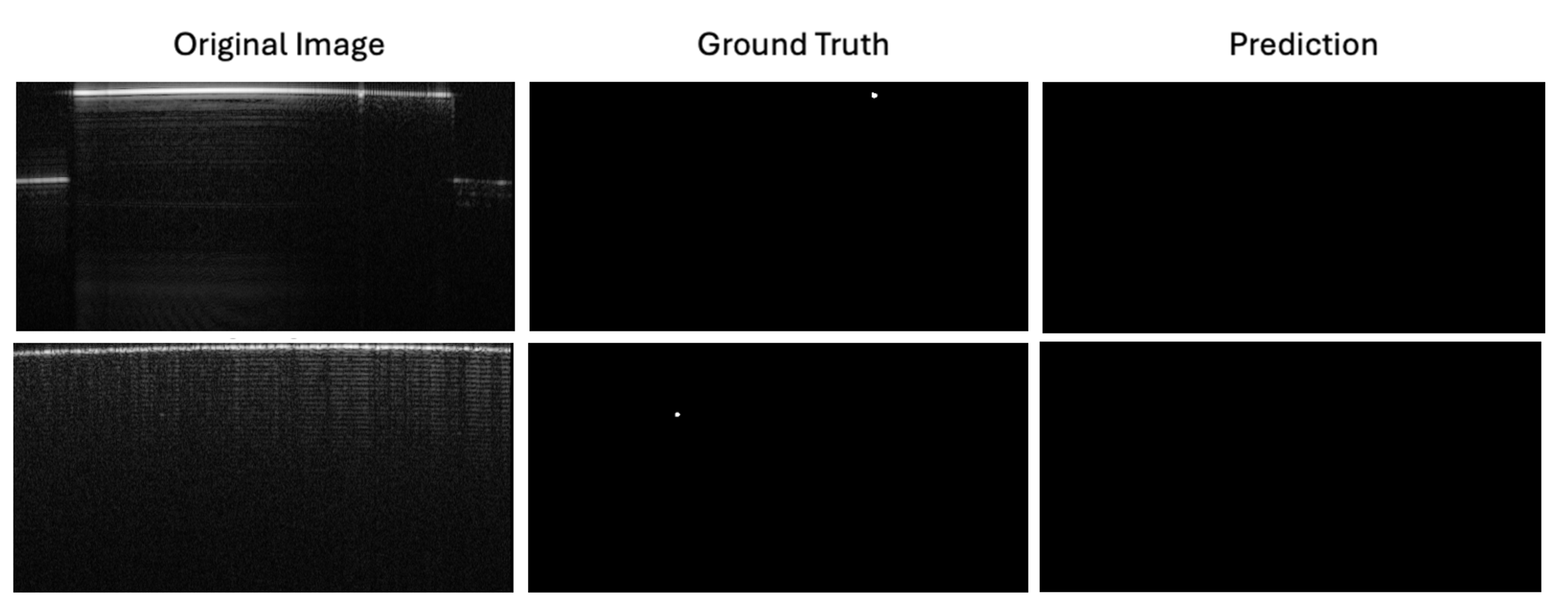}
\caption{Examples of a segmentation outcome (no Prediction) where the model did not capture the annotated defect, illustrating a failure case.}
\label{fig:cuali_bad}
\end{figure}

Figure~\ref{fig:cuali} showcases representative segmentation results in comparison with manual annotations. The model demonstrates high accuracy in detecting and localizing defects, with outputs that closely match expert labels. These results provide strong evidence of the effectiveness and reliability of the proposed approach. Nonetheless, as shown in Figure~\ref{fig:cuali_bad}, occasional failure cases occur, where certain defects remain undetected. Such instances, however, are relatively infrequent and highlight opportunities for further refinement.

\section{Conclusion}
This work demonstrates the successful application of a U-Net architecture for automatic defect segmentation in OCT scans of ceramic parts. The selected model achieved high accuracy, with a DSC of 0.979, outperforming comparable studies. The efficient inference time of 18.98 seconds per volume supports its viability for industrial quality control. The combination of precise segmentation and fast processing enables reliable automatic defect detection, with the potential to reduce inspection costs and improve manufacturing standards. 

Despite the good results of the trained model, this study has some important limitations. First, the dataset is relatively small, with only 30 volumes, which may limit the model’s ability to generalize to diverse situations. Second, the analysis focused exclusively on inclusion defects, as a preliminary step toward evaluating the feasibility of automated defect detection, while extending the approach to additional defect types remains an opportunity for future work.

\section*{Funding}

This project has received funding from Horizon Europe, the European Union’s Framework Programme for Research and Innovation, under Grant Agreement No. 101057404 (ZDZW), and Grant Agreement No. 101058054 (TURBO). This project has received funding from Villum Fonden (2021 Villum Investigator project No. 00037822: Table-Top Synchrotrons). Generalitat Valenciana partially funded this work through the project CIPROM/2022/20. The work of NPG was supported by the grant PID2022-140189OB-C21 funded by MICIU/AEI/10.13039/501100011033 ERDF/UE and FSE+.

%
%
%
\bibliographystyle{splncs04}
\bibliography{bibliography}

%




\end{document}